\documentclass[conference]{IEEEtran}
\IEEEoverridecommandlockouts
\usepackage{cite}
\usepackage{amsmath,amssymb,amsfonts}
\usepackage{algorithmic}
\usepackage{graphicx}
\usepackage{textcomp}
\usepackage{xcolor}
\usepackage{multirow}
\def\BibTeX{{\rm B\kern-.05em{\sc i\kern-.025em b}\kern-.08em
    T\kern-.1667em\lower.7ex\hbox{E}\kern-.125emX}}

\begin{document}


\title{Personality Type Based on Myers-Briggs Type Indicator with Text Posting Style by using Traditional and Deep Learning
\\
}

\author{\IEEEauthorblockN{Sakdipat Ontoum\IEEEauthorrefmark{1}\IEEEauthorrefmark{2},
Jonathan H. Chan\IEEEauthorrefmark{1}\IEEEauthorrefmark{2}}
\IEEEauthorblockA{\IEEEauthorrefmark{1}Computer Science Program, School of Information Technology}
\IEEEauthorblockA{\IEEEauthorrefmark{2}Innovative Cognitive Computing (IC2) Research Center} 
\IEEEauthorblockA{King Mongkut’s University of Technology Thonburi}
\IEEEauthorblockA{Bangmod, Thung-Khru, Bangkok, Thailand}

}

\maketitle

\begin{abstract}
The term “personality” may be expressed in terms of the individual differences in characteristics pattern of thinking, feeling, and behavior. This work presents several machine learning techniques including “Naive Bayes”, “Support Vector Machines”, and “Recurrent Neural Networks” to predict people personality from text based on “Myers-Briggs Type Indicator (MBTI)”. Furthermore, this project applies “CRISP-DM”, which stands for “Cross-Industry Standard Process for Data Mining”, to guide the learning process. Since, “CRISP-DM” is kind of iterative development, we have adopted it with agile methodology, which is a rapid iterative software development method, in order to reduce the development cycle to be minimal.
\end{abstract}


\section{Introduction}

The “Myers-Briggs Type Indicator (MBTI)” is an approach that reflects and identifies a person’s character. According to [2], a person’s personality is described as a group of characteristics that describe the likelihood of a person’s unique behavior, feelings, and thoughts. These characteristics of a person evolve and may differ in different contexts. In other words, personality is a collection of the traits that contribute to an individual’s distinct personality. Nowadays, many different types of personality models are utilized to characterize personalities, such as the “Big Five Personality Traits model” [3], “VIA Classification of Character Strengths” [4], “Myers-Briggs Type Indicator (MBTI)” model [5,6], and “Jung's Theory of Personality Type models”. It has been discovered that  “MBTI” is more powerful because it has a broader application in a variety of disciplines. However, it has a few reliability and validity concerns. We chose the  “MBTI” personality model for this study because of its popularity and the potential for utilizing it in a variety of fields.

Machine learning  is a method that researchers frequently employ to predict personality. They can utilize machine learning to learn personality patterns because it can be utilized to learn historical data and predict future data [1, 30]. In psychology science, such an application exists as a tool for personality evaluation and prediction [30]. The development of machine learning algorithm-based personality prediction technologies has become more important to businesses and recruiters for choosing the best candidates [30].

Nowadays, both natural language processing and social science communities have expressed an increasing interest in automatic personality prediction from social networks [9,30]. Social networks can also be utilized to investigate people’s personalities. A social networking post can be made in a variety of ways, including by utilizing an image, a URL link, and music. Unlike traditional personality tests, which have mostly been utilized in human resources management, counseling, and clinical psychology. The application of automatic personality prediction from social networks is much larger. Dating websites and social media marketing are two examples [10, 31]. 

In this paper, personality type prediction from the text by utilizing the “Myers-Briggs Type Indicator” on the  “MBTI” dataset to predict candidates' personality types [12]. The proposed method employs machine learning to create a classifier that utilizes text as input to predict the author's  “MBTI” personality type. The remainder of this paper is structured as follows: Section 2 discusses related works; Section 3 outlines the research methodology; Section 4 presents results and discussion; and Section 5 concludes the work.

\section{related works}

\subsection{“Machine Learning”}

“Machine learning” is used to train machines on effective handling of data. Sometimes it is difficult for human to interpret a pattern or extract information from vast amount of data. Machine learning is utilized in this case [13]. Generally, machine learning is grouped into three types. For instance, “supervised learning”, “unsupervised learning”, and “reinforcement learning” [32,33]. “Supervised learning” and “Unsupervised learning” are the most widely utilized and widely accepted methods [32]. The supervised machine learning algorithms are those that require assistance from a human. The training and testing datasets are separate, without overlap between them. An output variable must be predicted or classified in the training dataset. It is expected that an algorithm would learn patterns from training data and then apply them to test data to predict or classify. “K-Nearest Neighbor (KNN)”, “Logistic Regression”, and “Stochastic Gradient Descent” are examples of classification algorithms. 

Furthermore, unsupervised learning algorithms only extract a few features from the data. Then when unseen data is introduced, it utilizes previously learned features to predict the class of the new data. “K-Means”, “Mean Shift”, and “K models” are just a few examples [14]. Meanwhile, “reinforcement learning” is employed when the task at hand entails making a series of decisions that lead to a final reward. An artificial agent is rewarded or punished for the actions it takes during the training process. The goal of the learning process is to maximize total reward. “Q-Learning” and the “Markov Decision Process” are two examples of reinforcement learning algorithms. 

\subsection{“Personality Prediction System from Facebook Users” [10]}

“Facebook” has utilized Personality Prediction Systems for many years to predict a user's personality based on their “Facebook” functionalities [10,30,31]. For users' personalities, ”Facebook” utilized the “Big Five Personality Traits model” [31]. This model would be utilized to discover “Conscientiousness”, “Extraversion”, “Agreeableness”, “Openness”, and “Neuroticism” [10]. The researchers utilized two dataset collections in this study to predict the personalities of the users [30]. The first dataset is made up of data samples from the “myPersonality” project, while the second was built by hand [30]. Before proceeding to the next stage, English-language texts are adjusted in the preprocessing stage. The removal of URLs, symbols, names, and spaces, as well as the lowering of the case, stemming, and removal of stop words, are all part of the preprocessing steps. Slang and non-standard words in Bahasa Melayu data are replaced manually in a separate preprocessing phase before those of the texts are translated to English.

In this study, traditional machine learning algorithms and deep learning were utilized to run a series of tests to predict candidates' personality types for specifying job positions with maximum accuracy in the classification process. Traditional machine learning algorithms include “Support Vector Machines”, “Gradient Boosting”, “Naive Bayes”, “Linear Discriminant Analysis (LDA)”, and “Logistic Regression” [30]. Deep learning implementations, on the other hand, typically employ four architectures: “Multi-Layer Perceptrons (MLP)”, “Long Short-Term Memory (LSTM)”, and “1-Dimensional Convolutional Neural Networks (1D CNN)” [30].

Experimental algorithms of “deep learning” indicated that the “MLP architecture” had the highest average accuracy in the “myPersonality” dataset [10,30], while the “1D CNN+LSTM” architectures had the highest accuracy in the gathered dataset [10,30]. To summarize, “deep learning” algorithms can be utilized to improve dataset accuracy, even for low-accuracy traits. 

\subsection{“Myers–Briggs Type Indicator (MBTI)”}

Research in the personality field has long piqued the interest of psychologists, and one such study was conducted on the “Myers–Briggs Type Indicator” by a psychiatrist named “Carl Jung”. Then, “Katharine Briggs” and “Isabel Myers Briggs” created the “Myers-Briggs Type Indicator” for testing personality in the 1920s, based on “Jung's theory of psychological types” [5,8,30]. This model instrument has 16 personality types represented by a “personality types key” as shown in Fig. 1 [7]. In the “MBTI” system, for example, people classified as “INTPs” prefer “Introversion (I)”, “Intuition (N)”, “Thinking (T)”, and “Perception (P)” personality traits. We can classify the needs or behaviors of individuals according to labels, and then the machine can learn the patterns.

As shown in Fig. 2, the preferences in four dimensions are indicated by combining the 16 personality types. Each dimension corresponds to two distinct personalities. The four dimensions are “Introversion (I) – Extraversion (E)”, “Intuition (N) – Sensation (S)”, “Feeling (F) – Thinking (T)”, and “Perception (P) – Judgment (J)” [8].

Text or related data can be utilized to extract a variety of personality-related features. We can determine a user's  “MBTI” type by analyzing their posts on the social platform utilizing “Term Frequency-Inverse Document Frequency (TF-IDF)” to detect and quantify a person's most frequently utilized words. 

\begin{figure}[htbp]
\centerline{\includegraphics[scale=0.45]{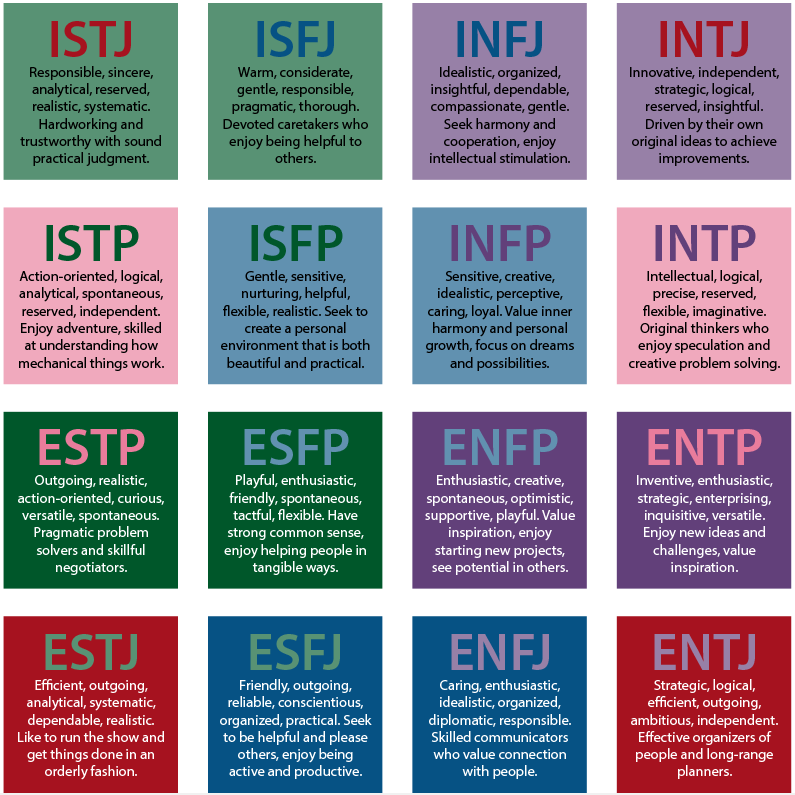}}
\caption{A chart with descriptions of each Myers–Briggs personality type [24]}
\end{figure}

\begin{figure}[htbp]
\centerline{\includegraphics[scale=0.45]{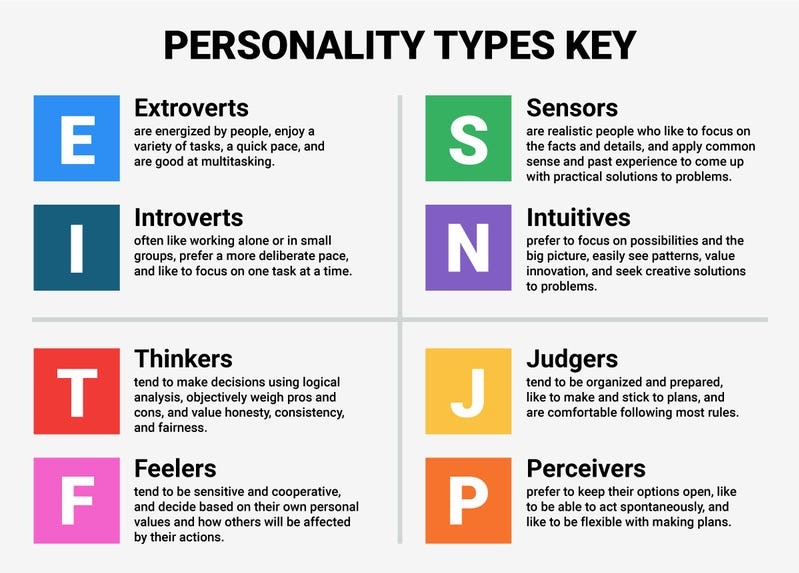}}
\caption{ Key Personality Types [25]}
\end{figure}

\subsection{“Recurrent Neural Networks” }

“Recurrent Neural Networks” are a subset of feed-forward neural networks that can send information along with time steps. They are a diverse family of models that can perform nearly arbitrary computations [19,20]. It has been demonstrated that finite-sized “Recurrent Neural Networks” with sigmoidal activation functions can simulate a universal Turing machine in a well-known result [20]. In practice, the ability to model temporal dependencies makes “Recurrent Neural Networks” particularly well suited to tasks where the input and/or output consists of non-independent sequences of points [20,21].

\textbf{Advantages}

\begin{itemize}
\item

“Recurrent Neutral Networks” can store records in such a way that each pattern is assumed to be dependent on previous ones.

\item

“Recurrent Neural Networks” are also utilized in combination with convolutional layers to broaden the effective pixel neighborhood [41][42].

\end{itemize}

\textbf{Disadvantages}

\begin{itemize}
\item

“Recurrent Neutral Networks” mostly have problems with exploding and vanishing gradients.

\item

It is hard to train “Recurrent Neutral Networks”.

\item

If tanh or ReLU is utilized as an activation function, it cannot handle very long sequences.
\end{itemize}

\subsection{“Support Vector Machine” Classifiers }

The “Support Vector Machines” Classifier is a supervised machine learning algorithm useful for “classification” and “regression tasks” [17,34,36]. However, it is mostly utilized in classification tasks [36]. Each data point is represented in N-dimensional space by the “Support Vector Machines” algorithm, and each feature is represented in a specific coordinate [36]. Then, it can classify the data by locating the hyperplane that best separates the two classes [18,27,34].  

\textbf{Advantages}

\begin{itemize}
\item

When there is a clear margin between classes, the “Support Vector Machines” perform reasonably well [35,39].

\item

In high-dimensional spaces, “Support Vector Machines” are more effective.

\item

“Support Vector Machines” are effective when the number of dimensions exceeds the number of samples [35,39].

\item

“Support Vector Machines” utilize a small amount of memory.

\end{itemize}

\textbf{Disadvantages}

\begin{itemize}
\item

The “Support Vector Machines” algorithm usually perform poorly for large datasets.

\item

If the dataset contains noise, overlap classes, and “Support Vector Machines” do not perform well [35].

\item

The “Support Vector Machines” may perform poorly when the number of features in each data point outstrips the sample size of the training data [35].

\end{itemize}

\subsection{“Naive Bayes” Classifiers }

The Bayes theorem is the foundation of the supervised machine learning algorithm on “Naive Bayes”. Each feature will contribute independently and equally to the target class or label [15,32]. Furthermore, when the likelihood of a sample belonging to a specific class is increased, Then they stop interacting with one another [32]. The “Naive Bayes” classifier is easy to implement, computationally fast, and effective on large, high-dimensional datasets [16,26,32]. 

\textbf{Advantages}

\begin{itemize}
\item
“Naive Bayes” can be used to solve multi-class prediction problems quickly and efficiently.

\item
When the feature independence hypothesis is true, it can outperform other models while requiring substantially less training data. 

\end{itemize}

\textbf{Disadvantages}

\begin{itemize}
\item

“Naive Bayes” presupposes that all characteristics are independent, which is seldom the case. This restricts the algorithm's usability in real-world use applications [40].

\item

This approach encounters the “zero-frequency problem” in which it assigns zero probability to a categorical variable whose category is in the test data set [40]. Also, it was not included in the training dataset [40].

\item

Because its predictions might be inaccurate in some instances, its probability outputs should not be taken too seriously. 
\end{itemize}

\section{Research Methodology}

\subsection{“CRoss Industry Standard Process for Data Mining (CRISP-DM)”}

A schematic of “CRoss Industry Standard Process for Data Mining (CRISP-DM)” is shown in Fig. 3. It is a free model that has become the industry standard in data mining methodology [38]. Because of its industry and tool independence, “CRISP-DM” can give instructions for the structured execution of any project. All scheduled tasks are often divided into six separate periods [11, 38].

\begin{figure}[htbp]
\centerline{\includegraphics[scale=0.46]{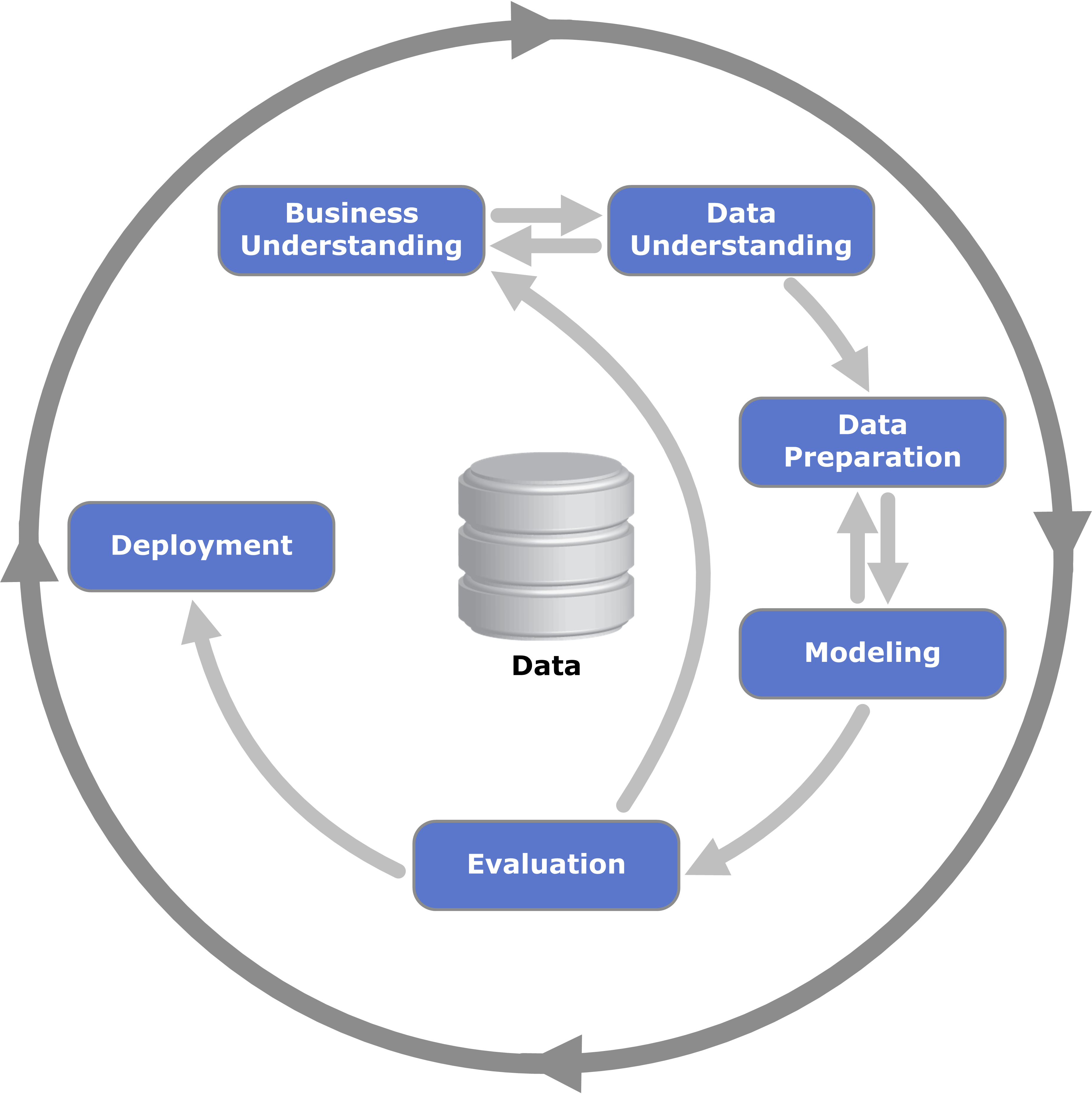}}
\caption{“CRISP-DM” process framework diagram [23]}
\end{figure}

\begin{itemize}
\item 

\textbf{“Business Understanding”:} Learn the domain, clarify the goal, and plan [22].

\item 

\textbf{“Data Understanding”: } Understand the data in all aspects [22]. 

\item 

\textbf{“Data Preparation”: } Preprocessing and feature engineering [22].

\item 

\textbf{“Modeling”: } Build the model [22].

\item 

\textbf{“Evaluation”: } Evaluate model performance [22].

\item 

\textbf{“Deployment”: } Deploy the model to end users [22].
\end{itemize}

The cycle process serves as a good overall framework for an end-to-end data science product [22]. However, when tied to a product cycle, particularly in the case of a modeling product, things become more complicated; this is where agile data science can come into play [22].

\subsection{“Agile Methodology” with “CRISP-DM”}

The agile development process is a development cycle that has been reduced to a bare minimum [22]. The key to agile data science is, to begin with, a minimal viable model as a baseline, then refine the model through multiple iterations. We may not build the perfect model during this process, but we will deliver a model with business value. In each sprint of our project, we utilize “Agile methodology” and “CRISP-DM” as shown in Fig. 4.

\begin{figure}[htbp]
\centerline{\includegraphics[scale=0.3]{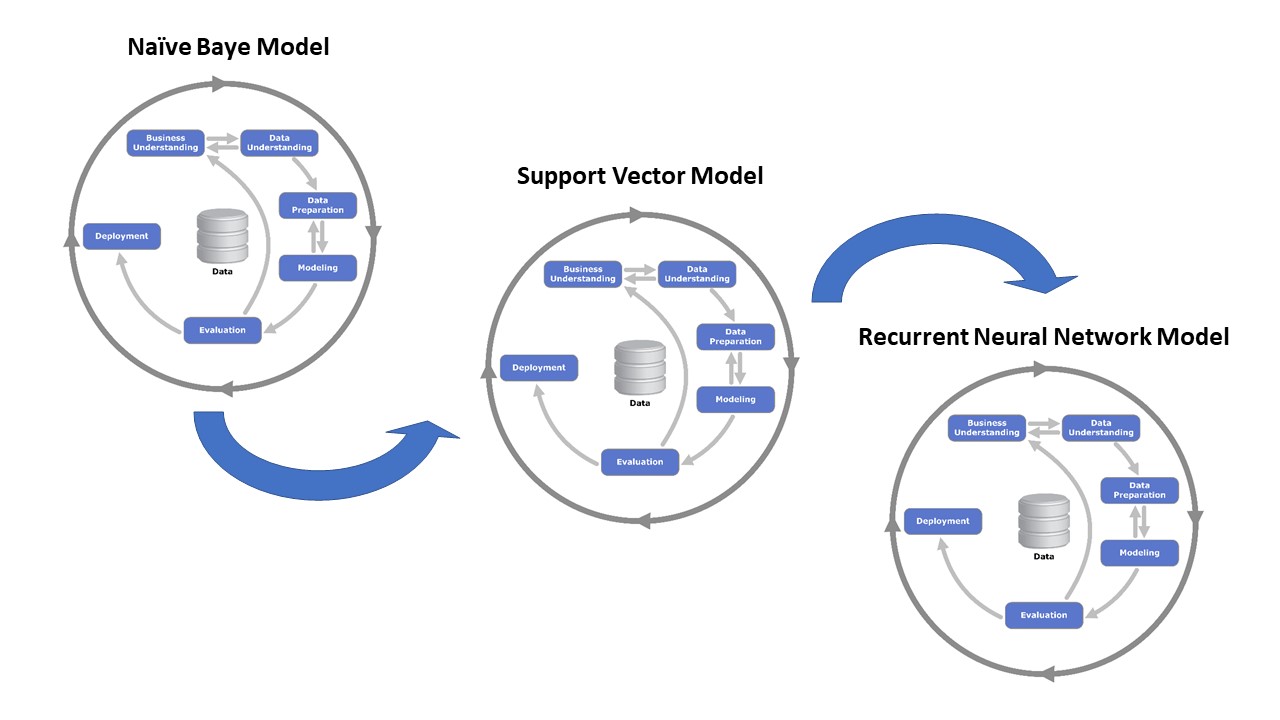}}
\caption{“CRISP-DM” process framework with Agile Development diagram}
\end{figure}

\subsection{Data Understanding}

\textbf{Dataset}

The  “MBTI” dataset is shown in Fig. 9, which is publicly available on “Kaggle” [12] and contains 8675 rows of post data. Each row has two columns, which are the personality type based on  “MBTI” and personal social networking posts from “personalitycafe.com”. Users first complete a questionnaire that determines their  “MBTI” type. Then, it allows them to publicly chat or forum with another user. since each user has fifty posts available to them. There are 430,000 data points in total. The data in each row contains the following:

\begin{itemize}
\item 

Type (This persons 4 letter MBTI code/type)
\item 

A section from each of their last 50 posts (each entry separated by “ $|$ $|$ $|$ ” (3 pipe characters)) [12].

\end{itemize}

\textbf{Exploratory Data Analysis}

The objective of the exploratory data analysis was to use various tables and graphs to generate a visual representation of the data for further investigation. As shown in Fig. 5, the proportionality of  “MBTI” type is shown for the general population. There is a significant imbalance in the “Introversion (I) - Extroversion (E)” and “Intuition (N) - Sensing (S)” pairs. On the other hand, “Perception (P) - Thinking (T)” and “Judgment (J)” pairs are quite balanced.

Then, we categorized the personality type keys into four dimensions as shown in Fig. 6. In the first category of “Extroversion (E) – Introversion (I)”, the distribution of “Extroversion (E)” is significantly bigger than that of “Introversion (I)”. In the second category, “Sensation (S) – Intuition (N)”, “Sensation (S)” have a significantly higher dispersion than “Intuition (N)”. In the third category, “Thinking (T) – Feeling (F)”, “Thinking (T)” has a somewhat greater distribution than “Feeling (F)”. Lastly, in the fourth category, “Judgment (J) – Perception (P)”, “Judgment (J)” is more prevalent than “Perception (P)”. 

Then, for the concepts that were used the most frequently by specific classes of the personality dimensions, we constructed word clouds. They were created by deleting postings that had the most extreme class probability prediction.
The size of each term in the word clouds is then proportional to its frequency of appearance in the top posts. We feel that these word clouds reflect some of the unique ways that different  “MBTI” utilize language, as shown in Figs. 7 and 8. 

\begin{figure*}[htbp]
\centerline{\includegraphics[scale=0.5]{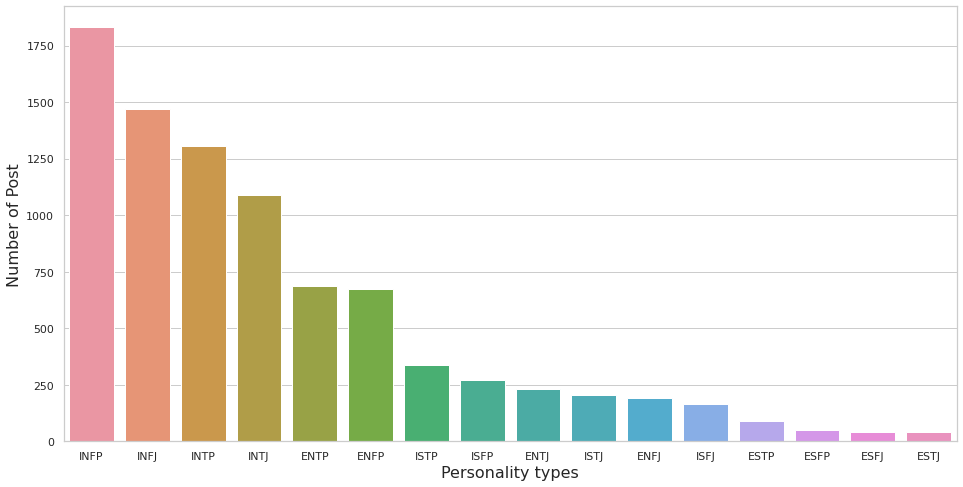}}
\caption{Proportionality diagram of  “MBTI” types}
\end{figure*}

\begin{figure*}[htbp]
\centerline{\includegraphics[scale=0.85]{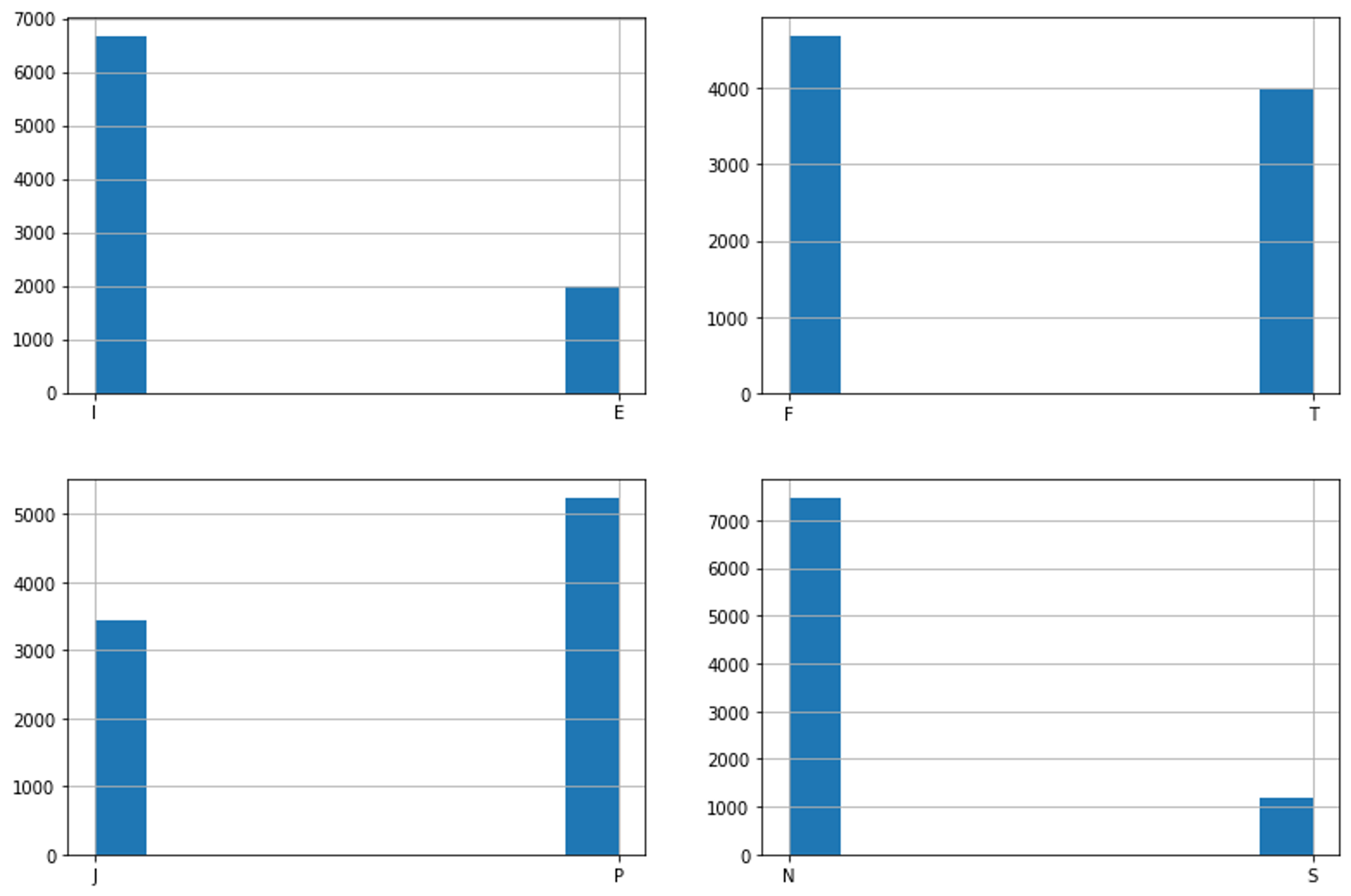}}
\caption{Proportionality diagram of each  “MBTI” type key}
\end{figure*}

\begin{figure*}[htbp]
\centerline{\includegraphics[scale=0.45]{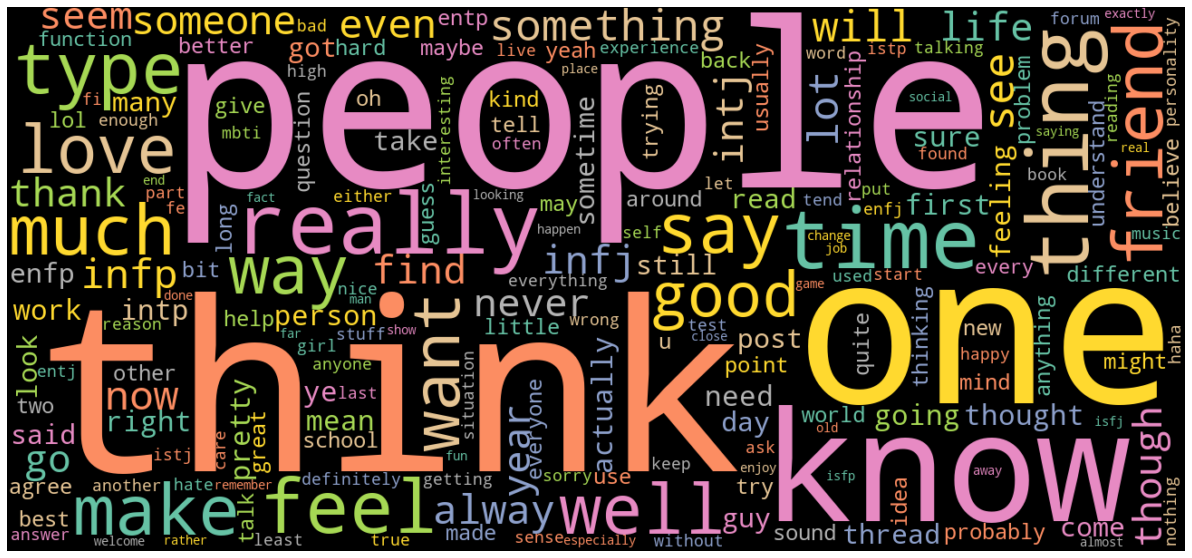}}
\caption{Unique word cloud for overall  “MBTI” types}
\end{figure*}

\begin{figure*}[htbp]
\begin{center}
\centerline{\includegraphics[scale=0.5]{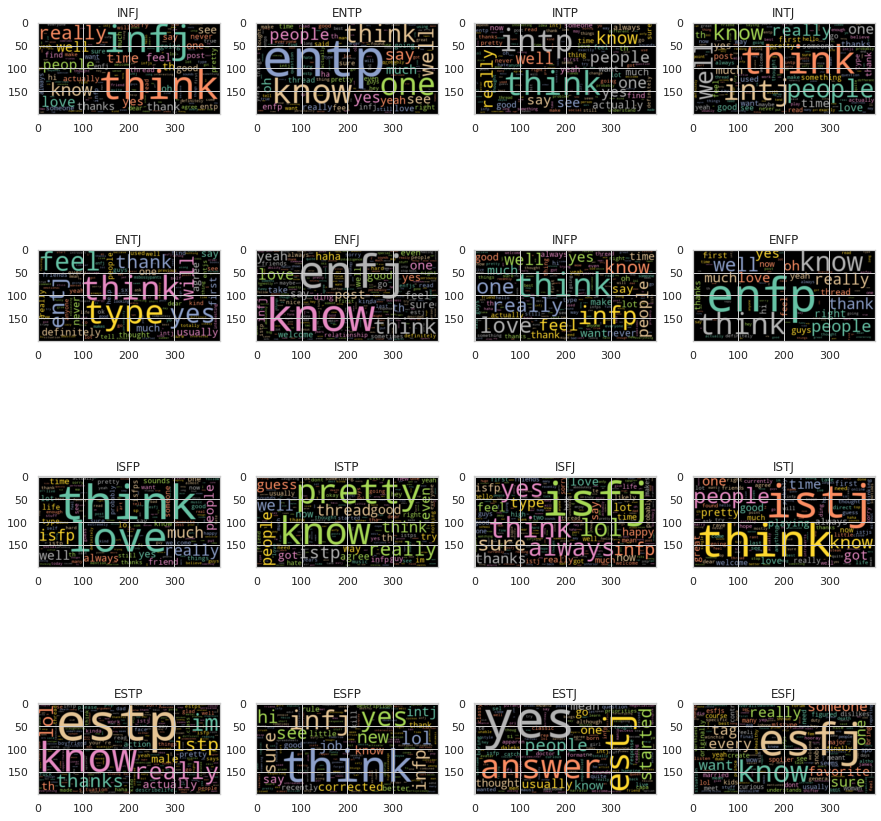}}
\end{center}
\caption{Unique word cloud for individual  “MBTI” types}
\end{figure*}

\begin{figure}[htbp]
\centerline{\includegraphics[scale=1.0]{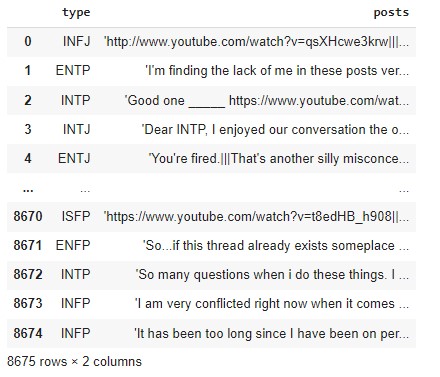}}
\caption{  “MBTI” Personality Type Dataset [12]}
\end{figure}

\subsection{Data Preparation}

\textbf{Preprocessing}

To better comprehend the datasets, we included four more columns that were categorized based on respondents' responses to the four dimensions of  “MBTI”: Introversion (I )- “Extraversion (E)”, “Intuition (N) - Sensation (S)”, “Feeling (F) - Thinking (T)”, and “Perception (P) - Judgment (J)”. The purpose of the procedure is to improve the accuracy of “Naive Bayes”, “Support Vector Machines”, and “Recurrent Neutral Networks”. When we added four columns on the four dimensions of “Recurrent Neutral Networks”, we utilized those variables as one-hot encodings with “pd.get\_dummies()”. values. 

\textbf{Selective Word and Character Removal}

Since the post data comes from a chat/forum named “personalitycafe.com” where people communicate solely through written text. We removed some data points that contained links to websites because we wanted our model to be generalized in the English language. Also, we utilized “Python's NLTK (Natural Language Toolkit)” to remove so-called “stop words” from the text. “NLTK (Natural Language Toolkit)” is a Python package for natural language processing.

\textbf{Lemmatization}

We will look at how inflected forms of the root word are transformed into dictionary forms. To lemmatize the text, we utilized “nltk.stem.wordNetLemmatizer”. This allows us to take advantage of the fact that inflections still have a single shared meaning.

\textbf{Tokenization}

\textbf{Tokenization for “Naive Bayes” and Support Vector Machiness Classifier}

We tokenized the words that had been transformed in the lemmatization process by utilizing an “NLTK word tokenizer”. That is, the common word is divided into small fractions of text. Then, we change the text to frequency by a “Bag of the Word (BoW)” and “Term Frequency-Inverse Document Frequency (TF-IDF)” to examine the relevance of key-words to documents in the corpus [28,37]. 

\textbf{Tokenization for “Recurrent Neutral Networks”}

We tokenized the words that had been transformed in the lemmatization process by utilizing a “Keras word tokenizer”. That is, the common word will change to be in position 1, position 2, and so on until 74,870 in total, respectively. Any other words that have been removed will now be in the form of lists of integers with a vocabulary of 1–74,870. Then, we change texts to sequences with 72,000 number words and pad the sequence with a max length of 200.

\textbf{Data Splitting}

To assess the correctness of the  “MBTI” personality model, the dataset was separated into two parts: training and testing. Utilizing the “train-test split” function in the “Scikit-learn” package, we segregated 75\% of the data for training and 25\% for testing. The testing dataset is a collection of previously unknown data that is solely utilized to evaluate the efficiency of a specific desired classifier.

\subsection{Modeling}
\begin{figure}[htbp]
\centerline{\includegraphics[scale=0.35]{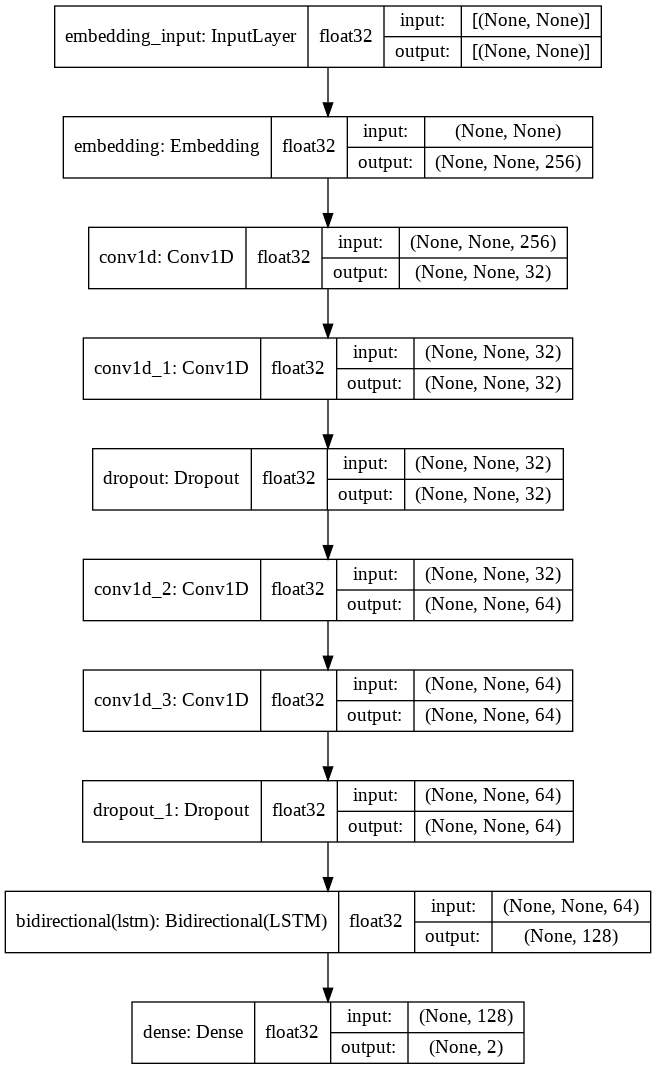}}
\caption{“Recurrent Neural Networks” Architecture}
\end{figure}

\begin{figure}[htbp]
\centerline{\includegraphics[scale=0.4]{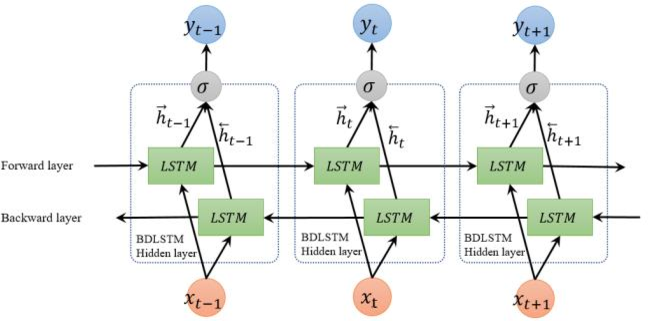}}
\caption{Bi-directional Long Short-Term Memory (BI-LSTM) Architecture [29]}
\end{figure}

\textbf{Classification Tasks}

We can divide the classification task from 16 classes into binary classes. Each binary class represents a unique dimension of personality as theorized by the “Myers-Briggs Type Indicator”. 

In terms of the “Naive Bayes” and “Support Vector Machines”, we imported the “Scikit-learn” library to generate those. Also, we utilized TensorFlow to create “Recurrent Neural Networks”. Then, utilizing the “Scikit-learn“ library's train-test split” function, the datasets were divided into “75\% training” and “25\% testing sets” for making a prediction, and the “Myers-Briggs Type Indicator” was trained individually. To begin with, we remove all columns that are unrelated to our features. 

\textbf{Modelling on “Naive Bayes” and “Support Vector Machines” classifiers}

We utilized “Scikit-learn” to construct the “Naive Bayes” classifier, which is a “multinomial-NB probabilistic” learning method that is commonly utilized in “Natural Language Processing (NLP)”. It computes the likelihood of each tag for a given sample and outputs the tag with the highest likelihood.

Second, we utilized “Support Vector Machines” with regularization parameter with cost with 1, kernel parameter is linear with degree 3, and we also applied gamma with auto to avoid exact match as per the training data set, which tends to cause over-fitting.

\textbf{Modelling on “Recurrent Neural Networks”}

As shown in Fig. 10, we utilize the embedding word to define the indexes into dense vectors of fixed size with a total length of vocabulary of 256. We decided to utilize “CONV1D” on “Recurrent Neural Networks” because “CONV1D” moves along a single axis. It makes perfect sense to apply this type of convolution layer to sequential data, such as text. Then, as shown in Fig. 11, to improve the performance of the “Recurrent Neural Networks” model, we added “Bi-directional Long Short-Term Memory (BI-LSTM)” with a size of 64, allowing the “Recurrent Neural Networks” to store sequence information in both directions, backward (future to past) and forward (past to future).

\section{Results and Discussion}

\


\begin{figure}[htbp]
\centerline{\includegraphics[scale=0.6]{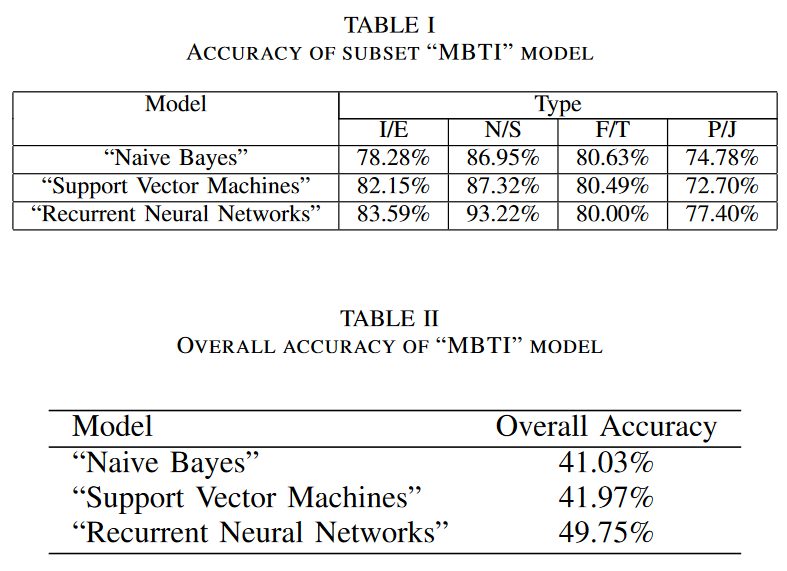}}
\end{figure}

Table I shows that the “Recurrent Neural Networks” outperform other machine learning models in terms of accuracy in all four personality types. The accuracy of “Recurrent Neural Networks” is significantly higher than that of “Naive Bayes” and “Support Vector Machines” for all categories. On the other hand, the accuracy of “Naive Bayes” and “Support Vector Machines” for “Perception (P)” and “Judgment (J)” is significantly lower than that of the “Recurrent Neural Networks”. As a result, the “Recurrent Neural Networks” outperform the other three machine learning models on this dataset.

In contrast with the overall accuracy in each model shown in Table II, there appears to be a general weakness in our model's ability to classify all four  “MBTI” dimensions accurately. However, the number that represents perfect classification does not indicate the efficacy of our model's predictions of overall  “MBTI” types.

Since the “Recurrent Neural Networks” outperform, the confusion matrix of the “Recurrent Neural Networks” model is shown in Fig. 12. For each  “MBTI” model, the results are mostly positioned as “True Positive” which means they are projected as positive and turn out to be true. which is a false positive, implying that the prediction is positive but incorrect.

\begin{figure}[htbp]
\centerline{\includegraphics[scale=0.4]{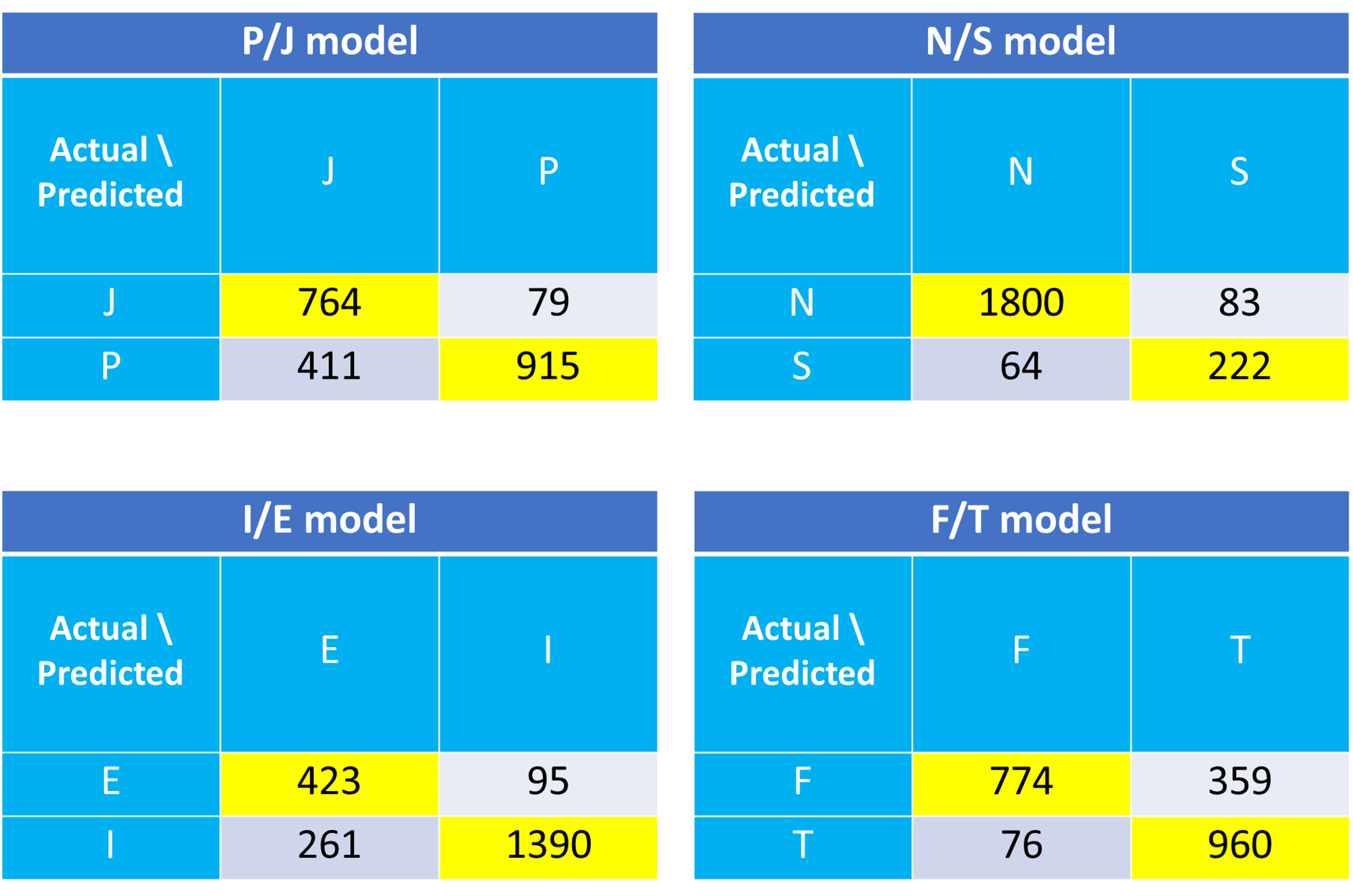}}
\caption{Confusion Matrix on each  “MBTI” models with RNN}
\end{figure}

\begin{figure}[htbp]
\centerline{\includegraphics[scale=0.5]{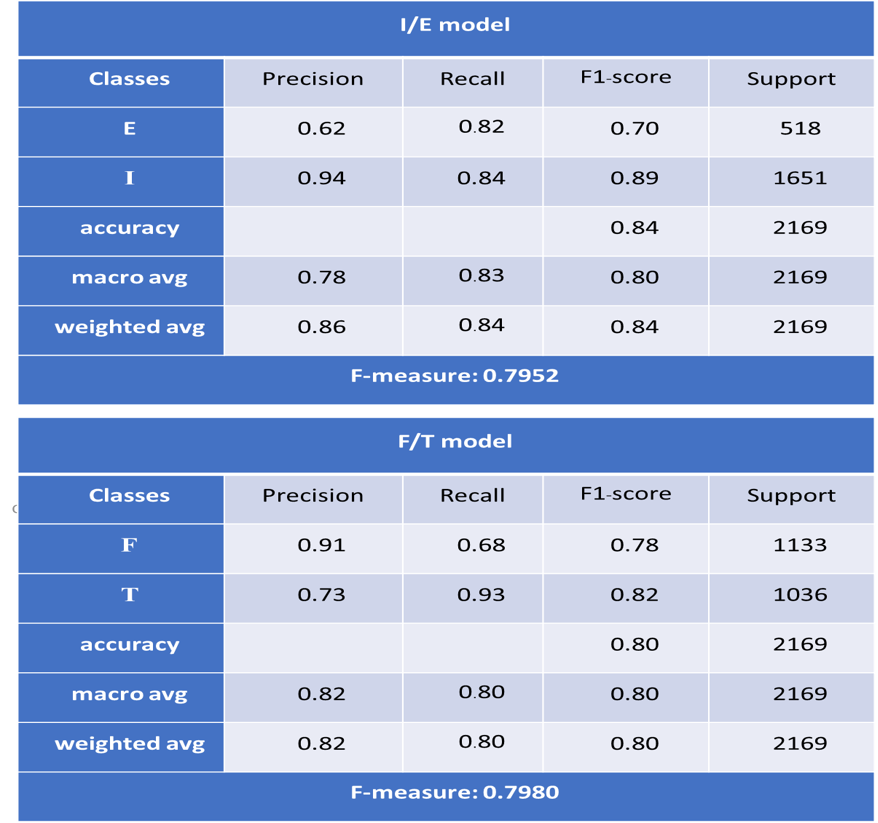}}
\caption{Classification Report on each  “MBTI” models with RNN (1)}
\end{figure}

\begin{figure}[htbp]
\centerline{\includegraphics[scale=0.5]{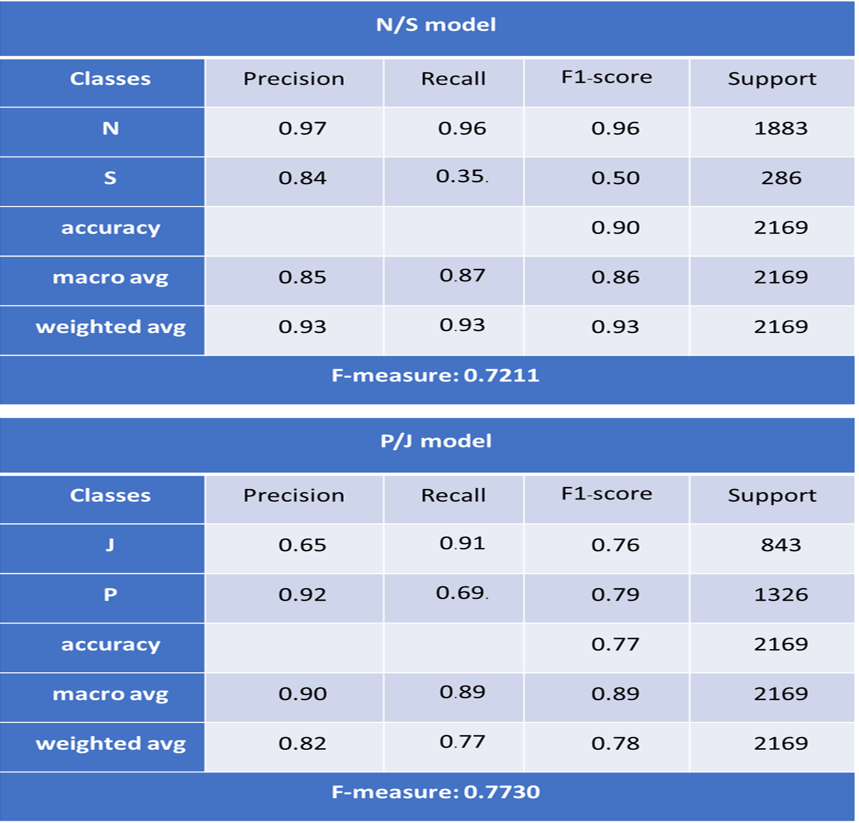}}
\caption{Classification Report on each  “MBTI” models with RNN (2)}
\end{figure}

Lastly, the classification report of the “Recurrent Neural Networks” is shown in Fig. 13 and Fig. 14. This shows that the F-score or F-measure, which is a weighted average score of the true positive (recall) and precision, is around 76\% when calculated as the mean of all models.  

\section{Conclusion}

In conclusion, this study was able to predict personality utilizing social network data with a machine learning algorithm that we applied. Also, the “Recurrent Neutral Networks” machine learning algorithm is the best model for predicting personality based on  “MBTI”. Companies would substantially benefit from this since they would be able to study their candidates' social network profiles before choosing suitable employees. 

\textbf{Limitation}

The “Myers-Briggs Type Indicator” from the text is only the first tier in developing a personality type model. The only social network utilized for this project was the Personality Cafe forum. Other social networks may provide useful data, allowing the prediction model to be improved. When building teams to produce new software, play sports, or fight crime, a technical position is one of the most important factors to consider when joining a team [30]. In addition, people's soft skills, such as mindset and personality, as well as their living environment must be considered.

\textbf{Future work}

We intend to utilize the Generative Pre-trained Transformer 3 (GPT-3) which is an autoregressive language model that leverages deep learning to generate human-like text [19].

\end{document}